\documentclass[times, review, 10pt]{elsarticle}

\usepackage{amssymb}
\usepackage{amsmath}
\usepackage{subfloat}
\usepackage{subfig}  % 提供 \subfloat 命令
\usepackage{subcaption}  % 替代 subfig，语法更清晰
\usepackage{graphicx}
\usepackage{diagbox}
\usepackage{algorithm}
\usepackage{algorithmic}
\usepackage{amsmath}
\usepackage{amsfonts}
\usepackage{amssymb}
\usepackage{booktabs}
\usepackage{multirow}
\usepackage{color}
\usepackage{float}
\usepackage[figuresright]{rotating}

\begin{document}

\begin{frontmatter}



\title{IllumFlow: Illumination-Adaptive Low-Light Enhancement via Conditional Rectified Flow and Retinex Decomposition}
\author[author1]{Wenyang Wei}
\ead{wenywei@stu.xidian.edu.cn}

\author[author1]{Yang yang}
\ead{23071213330@stu.xidian.edu.cn}
            
\author[author1]{Xixi Jia\corref{cor1}}
\ead{hsijiaxidian@gmail.com}
\cortext[cor1]{Corresponding author.}

\author[author1]{Xiangchu Feng}
\ead{xcfeng@mail.xidian.edu.cn}

\author[author1]{Weiwei Wang}
\ead{wwwang@mail.xidian.edu.cn}

\author[author2]{Renzhen Wang}
\ead{rzwang@mail.xjtu.edu.cn}

\address[author1]{School of Mathematics and Statistics,
            Xidian University, 
            Xi'an,
            710126, 
            Shaanxi,
            China}
            
\address[author2]{School of Mathematics and Statistics,
           Xi'an Jiaotong University, 
            Xi'an,
            Shaanxi,
            China}
% \affiliation{organization={School of Mathematics and Statistics},
%             addressline={ Xidian University}, 
%             city={Xi'an},
%             postcode={710126}, 
%             state={Shaanxi},
%             country={China}}


% \author{} %% Author name

% \affiliation{organization={},%Department and Organization

%% Abstract
\begin{abstract}
We present IllumFlow, a novel framework that synergizes conditional Rectified Flow (CRF) with Retinex theory for low-light image enhancement (LLIE). Our model addresses low-light enhancement through separate optimization of illumination and reflectance components, effectively handling both lighting variations and noise. Specifically, we first decompose an input image into reflectance and illumination components following Retinex theory. To model the wide dynamic range of illumination variations in low-light images, we propose a conditional rectified flow framework that represents illumination changes as a continuous flow field. While complex noise primarily resides in the reflectance component, we introduce a denoising network, enhanced by flow-derived data augmentation, to remove reflectance noise and chromatic aberration while preserving color fidelity. IllumFlow enables precise illumination adaptation across lighting conditions while naturally supporting customizable brightness enhancement. Extensive experiments on low-light enhancement and exposure correction demonstrate superior quantitative and qualitative performance over existing methods.
\end{abstract}


\begin{highlights}
\item Achieve more robust enhancement under various illumination conditions.
\item Incorporate an enhanced denoising module with rectified flow-guided data augmentation.
\item Enable bidirectional illumination adjustment.
\end{highlights}

%% Keywords

\end{frontmatter}



\section{Introduction}
Images acquired under low-light conditions frequently suffer from significant noise, detail loss, chromatic distortion, and diminished contrast \cite{weng2024mamballie, zhou2024glare}. Low-light image enhancement (LLIE) techniques are designed to address these degradations, reconstructing perceptually coherent visuals while preserving structural and radiometric integrity, and it is a long-standing yet vital challenge in computer vision \cite{hou2023global}. Despite notable progress in LLIE, the inherent complexity of degradation - marked by image-specific illumination variations and intricate noise structures - continues to pose significant challenges for reliable restoration \cite{zhou2025low}.

Numerous studies have been conducted on low-light enhancement techniques, evolving from traditional model-based methods \cite{lee2013contrast,guo2016lime,  xu2022novel, gu2019novel, hao2020low, jeon2024low} to learning-based approaches \cite{yang2021sparse, shen2017msr, guo2020zero, wang2020lightening, wei2018deep, jiang2021enlightengan, jiang2024exposure, sheng2025low}. Recent years have seen significant breakthroughs in learning-based methods for low-light image enhancement (LLIE) \cite{zheng2021adaptive, wang2022low, wu2022uretinex, cai2023retinexformer, he2023reti, yi2023diff, 10244055}.

\begin{figure}[H]
	\centering
	\includegraphics[width=3.2in]{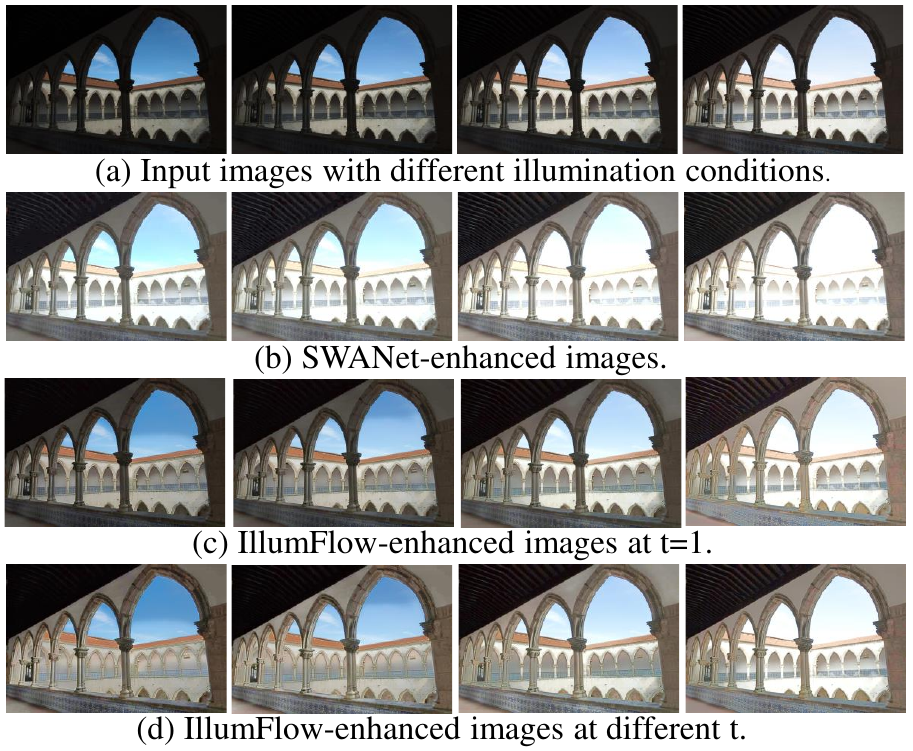}
    \\	
	\caption{Comparison of SWANet \cite{10244055} and our method outputs for input images under different lighting conditions on datasets\cite{afifi2021learning} originally rendered from the MIT-Adobe FiveK dataset.}
	\label{fig 00}
\end{figure}

\begin{figure}[H]
	\centering
	\includegraphics[width=2.8in]{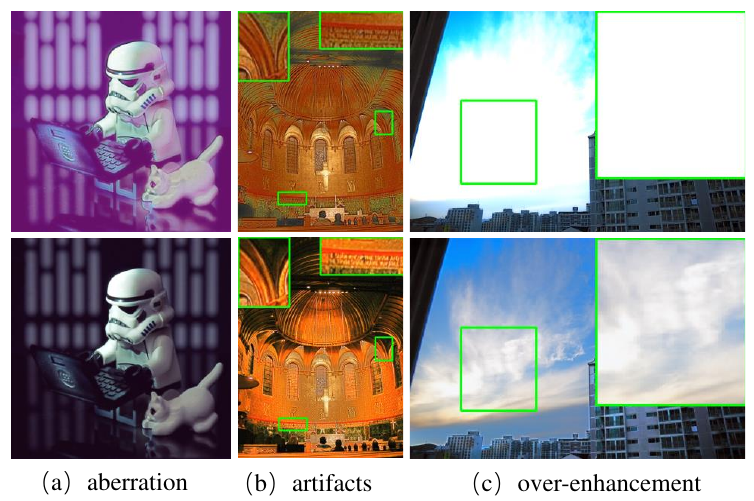}
    \\
    \includegraphics[width=3in]{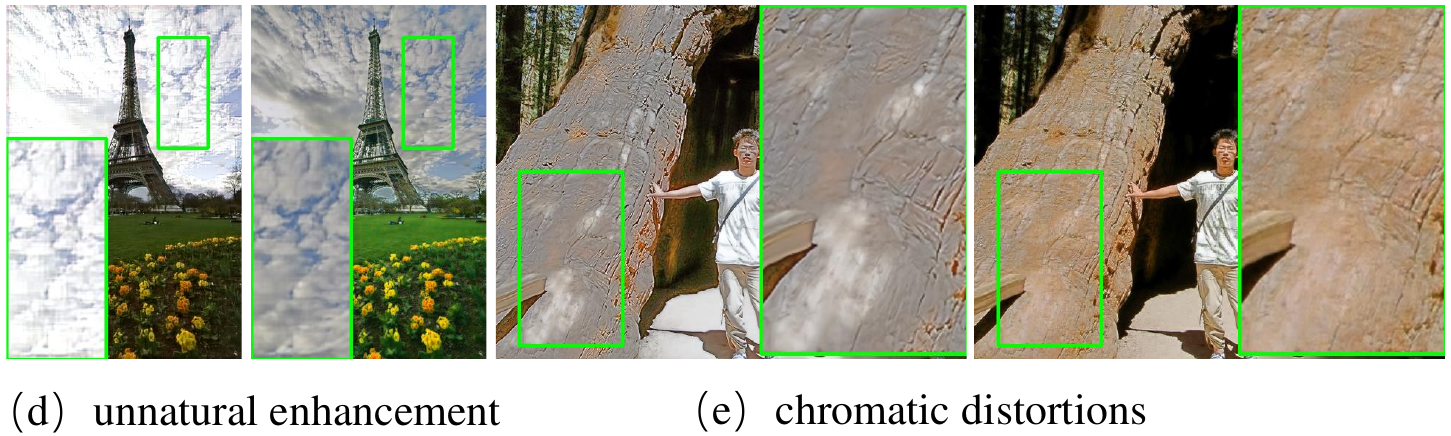}
	\caption{Comparison with learning- and diffusion-based methods: (a-c) RetinexNet  \cite{wei2018deep} shows artifacts/over-enhancement (top) vs. our method's cleaner results (bottom); (d-e) DiffLL \cite{jiang2023low} produces unnatural colors (left) while our method maintains better color fidelity and illumination (right).}
	\label{fig 0}
\end{figure}

Most of the learning-based approaches leverage end-to-end training to learn a mapping from low-light to natural images in a supervised manner. This learning framework predominantly depends on carefully curated training data \cite{yin2023cle}. However, a domain gap typically exists between real-world low-light images and the training data \cite{yang2023difflle}, causing learning-based methods to fail when generalizing to illumination conditions beyond the training distribution, as illustrated in Fig. \ref{fig 00}(a-b), where the SWANet fails to fully correct exposure and preserve color fidelity under wide lighting variations. This often results in suboptimal enhancement effects, including over-/under-exposure, amplified noise, and chromatic distortions, as evidenced in Fig. \ref{fig 0}(a-c) and reported by \cite{hou2023global, xue2024low}.

Recently, diffusion models \cite{ho2020denoising, song2020denoising} have emerged as a promising approach for LLIE, leveraging their exceptional ability to model natural image priors to significantly enhance existing enhancement frameworks. Unlike end-to-end training approaches, diffusion models address LLIE through a progressive process by learning a conditional image denoiser. Specifically, following \cite{yin2023cle, yi2023diff, jiang2023low, xue2024low, ooi2023llde, hou2023global}, diffusion-based LLIE methods progressively transform random noise into enhanced natural images through a conditional denoising process guided by the low-light input. These diffusion-based approaches consistently outperform conventional end-to-end trained methods, achieving state-of-the-art performance on benchmark datasets. 

For instance, DiffLL \cite{jiang2023low} introduces two key components: (1) a wavelet-conditional diffusion module to address color distortion and artifacts, and (2) a high-frequency reconstruction module for enhanced detail restoration. Diff-Retinex \cite{yi2023diff} proposes a dual-branch diffusion architecture grounded in Retinex theory, employing progressive denoising for the reflectance component while simultaneously enhancing the illumination layer. CLE-diffusion \cite{yin2023cle} achieves controllable illumination adjustment through explicit conditioning on lighting hyper-parameters, demonstrating preliminary yet promising light-manipulation capabilities.

While the diffusion process's strong generative capability helps address certain challenges in low-light enhancement, its performance remains highly sensitive to the conditional low-light input. Moreover, these methods often exhibit limited generalization capability when processing images with varying illumination intensities, leading to unnatural enhancement effects and chromatic distortions, as illustrated in Fig. \ref{fig 0}(d-e). Additionally, their long training time and slow inference speed remain major limitations.

The limited generalization capability of both end-to-end learning-based and diffusion-based LLIE methods stems from their reliance on constrained, carefully curated training datasets, such as the LOLv1 dataset. In real-world scenarios, however, degraded input images typically originate from diverse low-light conditions, resulting in substantial illumination variations that must be addressed during enhancement. Consequently, a significant distribution shift exists between the carefully curated training dataset and real-world low-light images.
 
In this paper, we demonstrate that illumination variations can be approximated by a linear parametric function. This linear relationship stems from the fundamental physical property that raw pixel values scale linearly with the radiant energy collected \cite{martinec2008noise,hasinoff2010noise}. This formulation enables efficient generation of realistic low-light images with diverse illumination patterns that closely approximate real-world conditions. By leveraging this diverse set of generated low-light images, we can train models that achieve significantly improved generalization to real-world low-light conditions. To accomplish this, we utilize a conditional rectified flow (CRF) model \cite{lipman2022flow} to process the illumination component. 

The CRF model learns an optimal ODE (Ordinary Differential Equation) governing the input-to-target transition by training on all intermediate interpolated data, ensuring robustness across the entire transformation. We observed that the illumination component of a low-light image and its corresponding natural counterpart follow a linear trajectory, making it well-suited for Conditional Rectified Flow (CRF) due to its inherent capacity for continuous, controllable adjustment. For the reflectance component of low-light images - which contains significantly more complex noise patterns than its natural-light counterpart - we employ a dedicated denoising model to enhance reflectance fidelity. We refer to this model as IllumFlow, and its robust performance under various illumination conditions is illustrated in Fig. \ref{fig 00} (c-d).

Specifically, our IllumFlow model begins by decomposing the input low-light image into illumination and reflectance components using a pretrained Retinex network. For illumination enhancement, we employ a conditional rectified flow model (CRFI - conditional rectified flow for illumination) to learn an continuous ODE flow field that characterizes the dynamic transformation from low-light to natural illumination conditions. Furthermore, the bidirectional ODE framework enables adaptive illumination mapping between arbitrary lighting conditions, facilitating customizable brightness enhancement.
Since complex noise is predominantly concentrated in the reflectance component, we propose a denoising network enhanced by flow-based data augmentation, which effectively removes reflectance noise and chromatic aberrations while maintaining color fidelity.
Unlike diffusion models that rely on a standard Gaussian as the initial distribution, our enhanced denoising module CRFR starts directly from the low-light reflectance distribution and reconstructs the normal-light reflectance through a one-step inference.

The main contributions of this paper are summarized as follows: 
\begin{itemize}
    \item Benefiting from the learning of continuous flow fields, our method achieves more robust enhancement under various illumination conditions compared to existing approaches.
    \item Our method incorporates an enhanced denoising module with rectified flow-guided data augmentation, enabling more efficient training and inference while maintaining superior noise removal and color fidelity.
    \item Our method enables bidirectional illumination adjustment. This capability expands the potential for other downstream applications, surpassing the limitations of diffusion-based approaches.
    \item Extensive experiments on low-light enhancement and related tasks validate the effectiveness of our method.
\end{itemize}

\section{Related  Work}
\subsection{Retinex Variational Model}

The effectiveness of Retinex decomposition in low-light image enhancement has been widely studied, with both hand-crafted and learned priors yielding promising results. According to Retinex theory \cite{land1977retinex, land1983recent}, an image can be decomposed into illumination and reflectance layers. Mathematically, the model can be described as 
\begin{equation}
	\label{equ1}
	I = L \odot R,
\end{equation}
where $I$ denotes the observed image, $L$ represents the illumination layer, $R$ is the reflectance layer, and $\odot$ indicates element-wise multiplication.

In Retinex variational frameworks, the illumination layer is constrained to be piecewise smooth, whereas the reflectance layer encodes the inherent scene characteristics. This theory has inspired multiple variational methods \cite{guo2016lime, xu2022novel, gu2019novel, hao2020low, cai2017joint, fu2019hybrid, ren2020lr3m} that perform low-light enhancement by incorporating different prior constraints for illumination and reflectance estimation. The standard variational framework for Retinex-based decomposition is given by:
\begin{equation}
	\label{equ2}
	{\arg \min }_{L,R} \; \phi(I,L,R) + {\lambda _1}{\psi _1}\left( L \right) + {\lambda _2}{\psi _2}\left( R \right),
\end{equation}
where $ \phi(I, L, R) $ represents the data fidelity term, while $\psi_1(L)$ and $\psi_2(R)$ denote the regularization terms for the illumination and reflectance layers respectively, with ${\lambda _1}, {\lambda_2} > 0$ being the regularization parameters that balance the contributions of the data fidelity term $\phi$ and the regularization terms $\psi_1, \psi_2$.

\subsection{Conditional Flow Matching and Rectified Flow}
Flow Matching \cite{lipman2022flow} has emerged as a flexible generative approach, capable of direct distribution-to-distribution mapping \cite{tong2023improving}. For low-light enhancement, this enables smooth and adaptive illumination adjustment without iterative refinement—addressing key limitations of diffusion-based methods. Flow Matching models a
probability path between distributions using an ODE. The process is determined by a vector field
$u_t(z):[0,1]\times \mathbb{R}^d \to \mathbb{R}^d$, which produces a flow $\varphi_t(x):[0,1]\times \mathbb{R}^d \to \mathbb{R}^d$ describing the evolution of samples. Formally, the flow satisfies:
\begin{equation}
	\label{equ3}
	\frac{{d{\varphi_t(x)}}}{{dt}} = {u_t}\left( {\varphi_t({x})} \right),
\end{equation}
where $\varphi_0(x) = x$ with $x$ drawn from the base distribution $p_0$ and the flow $\varphi_t(x)$ transforms the distribution $p_0$ into $p_1$ overtime, satisfying $\varphi_1(x) = y$ where $y$ is drawn from the target distribution $p_1$. Let $z_t$ denotes $\varphi_t(x)$, the time-dependent vector field $u_t(z_t)$ generates a probability path $q_t$ that evolves continuously from the initial distribution $p_0$ to the target distribution $p_1$. Once we have $u_t(\cdot)$, then given any initialization $\varphi_0$, we are able to obtain the $\varphi_t$ by solving the ODE in Eq. (\ref{equ3}).

In practice, the vector field $u_t(z_t)$ is approximated by a parameterized model $v_t(z_t;\theta)$, learned through the optimization of a flow matching objective that leverages conditional vector fields $u_t(z_t|\alpha)$ and their associated conditional probability paths $p_t(z_t|\alpha)$, which is called Conditional Flow Matching (CFM) \cite{lipman2022flow}. Specifically, the parameters $\theta$ are obtained by solving the following Eq. (\ref{equ5}):
\begin{equation}
	\label{equ5}
	{L_{CFM}}\left( \theta  \right) = {E_{t,{p_t}\left( {z_t|\alpha} \right),q\left(\alpha \right)}}{\left\| {{v_t}\left( {z_t;\theta} \right) - {u_t}\left( {z_t|\alpha} \right)} \right\|^2},
\end{equation}
where $q(\alpha)$ is density distribution over $\alpha$. 

The CFM enables the construction of different conditional probability vector fields and paths, including Variance Exploding (VE) \cite{song2020score}, Variance Preserving(VP) \cite{ho2020denoising} and conditional rectified flow (CRF) \cite{liu2022flow}. 
The conditional velocity field $u_t(z|\alpha)$ specifies the direction from an initial sample $x \sim p_0$ to its corresponding target sample $y \sim p_1$. The CRF produces straight line flow $\varphi_t(x)$ via the relation:
$\varphi_t(x) = (1-t)x +ty$ and
$u_t(z_t|\alpha) = y - x$,
where $\alpha = (x,y)$.

\section{The Proposed Method}
\begin{figure}[H]
	\centering
	\includegraphics[width=1\textwidth]{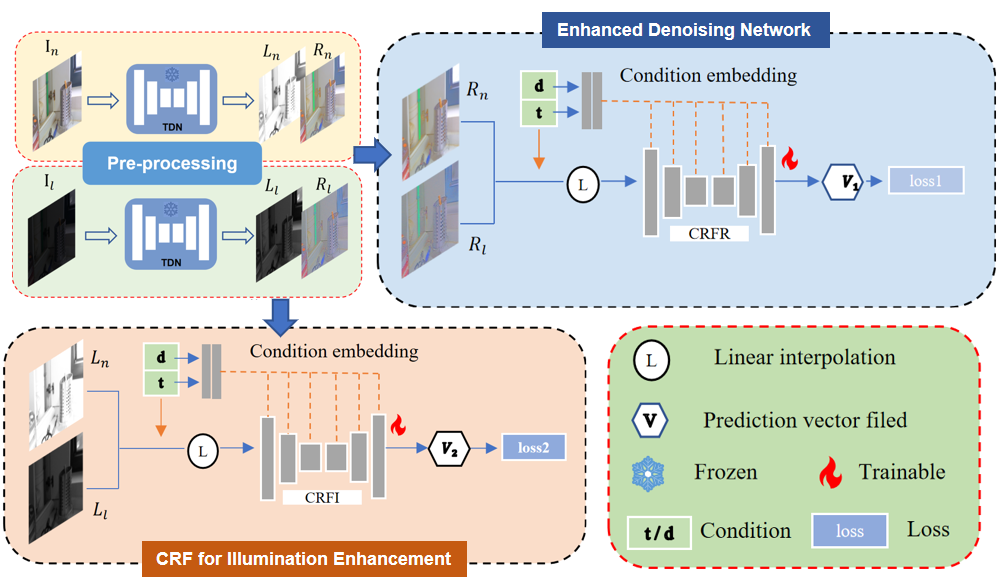}
	\caption{Training process. This framework involves: (1) TDN-based image decomposition to separate illumination/reflectance layers; (2) Enhanced denoising of reflectance; (3) Flow-based (CRF) continuous illumination enhancement for smooth brightness adjustment.}
	\label{fig1}
\end{figure}

In this section, we present IllumFlow, a low-light enhancement framework that adaptively optimizes illumination 
through a conditional rectified flow model integrated with Retinex decomposition. The IllumFlow framework comprises three key components, as shown in Figure \ref{fig1}. 

\subsection{Pretrained Retinex Model for Decomposition}
We employ a pretrained Transformer Decomposition Network (TDN) from Diff-Retinex \cite{yi2023diff} to decompose input images into illumination and reflectance components. Specifically, the TDN architecture processes both a normal-light image $I_n$ and a low-light image $I_l$, yielding two distinct representations for each:
\begin{itemize}
\item \textbf{Reflectance}: $R_n, R_l \in \mathbb{R}^{H \times W \times 3}$ (for normal and low-light conditions),
\item \textbf{Illumination}: $L_n, L_l \in \mathbb{R}^{H \times W \times 1}$ represent the scene’s illumination under each condition.
\end{itemize}
The illumination and reflectance components undergo distinct processing pipelines, which we elaborate on in the subsequent subsections.

\subsection{CRF for Illumination Enhancement}
\begin{figure}[!ht]
	\centering
 \includegraphics[width=1\textwidth]{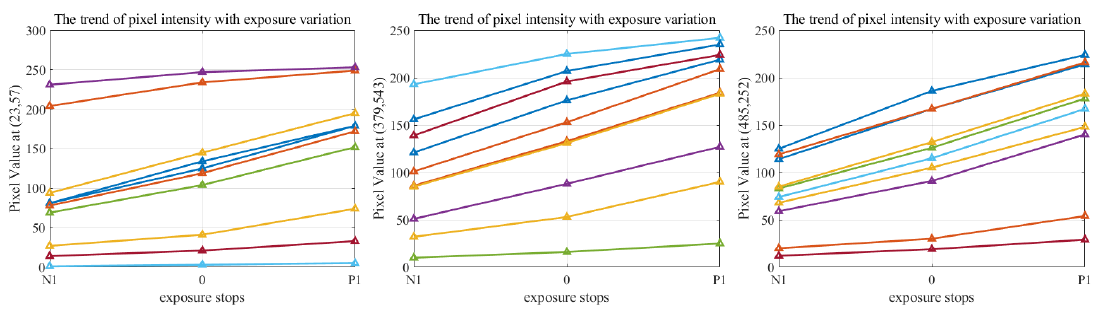}
	\caption{Variations in pixel values at fixed positions across ten representative images as a function of exposure time.}
	\label{fig 1-2}
\end{figure}

A key challenge in low-light enhancement is the significant variation in illumination conditions across different input images \cite{zhao2025ref}, as illustrated in Fig. \ref{fig 00}(a). 
These multi-exposure sequences of the same scene share a common irradiance $E$, and are modeled by Equ. \eqref{equ aa} as
\begin{equation}
\label{equ aa}
    I_{k} = f(E \cdot \delta_k),
\end{equation}
where $\delta_k$ is the exposure time, $f$ denotes camera response function and $I_{k}$ denotes the multi-exposures image sequences \cite{endo2017deep}. This variability demands adaptive processing to achieve robust enhancement results. 
In this paper, we demonstrate that illumination intensity variations follow an approximately linear relationship over a suitable exposure durations. In Fig. \ref{fig 1-2}, we illustrate the variations in pixel values at randomly selected fixed position $(23,57), (379,543), (485,252)$ across ten representative images, with each color corresponding to one of the ten images. The results show that the pixel value at this position exhibits a linear or piecewise linear dependence on the exposure time. 
Note that $N$ represents exposure reduction, $0$ represents normal exposure, $P$ represents exposure boost, and the number represents the ratio. 

Building on this observation, we propose to model the illumination transformation from low-light to normal-light conditions as a continuous flow $\varphi_t(x)$, enabling adaptive and smooth enhancement. Specifically, we employ conditional rectified flow to model $\varphi_t(x)$ as straight line flow
\begin{equation}
\label{equ 5-3}
    \varphi_t(x) = ty + (1-t) x, t\in [0,1],
\end{equation}
 where $x\sim P(L_l)$ and $y \sim P(L_n)$. Correspondingly, the conditional velocity field $u(z_t|\alpha)$ in Eq. (\ref{equ5}) is computed per-sample as the difference between a normal-light sample and a low-light sample, yielding $u(z_t|\alpha) = L_n - L_l$.

To achieve this, we propose a parameterized network module called CRFI (Conditional Rectified Field for Illumination Enhancement, Fig. \ref{fig1}) that learns to predict the vector field $v_t(\varphi_t(x);\theta)$ through the mapping:
\begin{equation}
\mathrm{CRFI_\theta}(\varphi_t(x),t,d) = v_t(\varphi_t(x);\theta).
\end{equation}
The network approximates the known conditional vector field $u(\varphi_t(x)|\alpha) = L_n - L_l$. The conditional rectified flow enables effective mapping from arbitrary intermediate illumination distributions $P(\varphi_t(x)), t\in [0,1]$ to the target normal-light distribution $P(L_n)$.  This is achieved by leveraging the relationship established in Eq. (\ref{equ 5-3}), which ensures that
\begin{equation}
\frac{d\varphi_t(x)}{dt} = u(z_t|\alpha) = L_n - L_l.
\end{equation}
For any intermediate state $\varphi_t(x)$ (representing illumination at varying intensity levels), the corresponding vector field $v_t(\varphi_t(x);\theta)$ serves as an approximation of $u(\varphi_t(x)|\alpha)$. Consequently, the final transformation can be expressed as follows:
\begin{equation}
\label{pht}
    \varphi_1(x) = \varphi_t(x) + \int_t^1 u(\varphi_s(x)|\alpha) ds,
\end{equation} 
where the integral accumulates the infinitesimal deformations governed by $u$. Thus, Eq. (\ref{pht}) demonstrates superior adaptability to diverse illumination variations in low-light images.

To optimize the network module, we employ a composite loss function combining the objective from Eq. (\ref{equ5}) with a consistency regularization term adopted from \cite{frans2024one}, defined as:
\begin{align}
    loss_1 =  {L_\mathrm{CRFI - C}} &= \left\| {{v_{t}(\varphi_t(x);\theta)} - \left( {L_n - L_l} \right)} \right\|_F^2  +\left\|  {v_{t}(\varphi_t(x);\theta,2d)}-{s_{\mathrm{target-L}}}  \right\|_F^2, \label{equ6} 
	% & +\left\| {\mathrm{CRFI}_{\theta_{2}} \left( {{L_t},t,2d} \right) - {s_{\mathrm{target-L}}}} \right\|_F^2), 
 \end{align}
 where predict vector filed $v_{t}(\varphi_t(x);\theta) = \mathrm{CRFI_\theta}(\varphi_t(x),t,d)$ with $d=0$ , 
 the consistency regularization term $s_\mathrm{target-L}
=\frac{v_t(\varphi_t(x);\theta,d)+v_t(\varphi_{t+d}(x^\prime); \theta, d)}{2}$ ,
$\varphi_{t+d}(x^{\prime}) = \varphi_t(x) + dv_t(\varphi_t(x);\theta,d)$ and $d$ denotes the time step. Note that $\varphi_{t+d}(x)$ is different from $\varphi_{t+d}(x^{\prime})$, the former is derived from Eq. (\ref{equ 5-3}), and the latter is derived from the Euler iteration.
By combining the conditional flow matching loss and consistency regularization loss from the shortcut model, the transition path from low-light illumination distribution to normal-light illumination distribution is straighter and smoother.

\begin{figure*}[ht]
	\centering
	\includegraphics[width=1\textwidth]{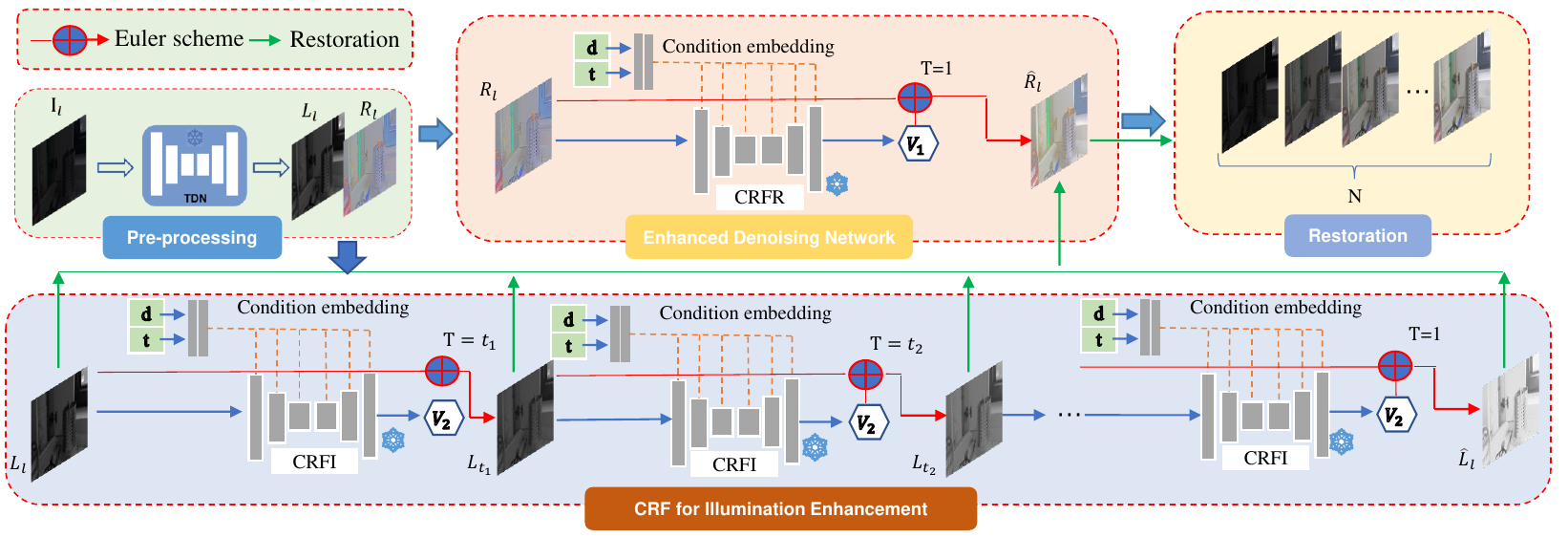}
	\caption{Inference process. Color-coded framework components: Green: TDN decomposition (preprocessing); Pink: Enhanced denoising module; Purple: Flow-based illumination enhancement (progressive process; single CRFI = one-step enhancement); Yellow: Final output (denoised reflectance + enhanced illumination sequence).}
	\label{fig 2}
\end{figure*}

\subsection{Enhanced Denoising Model for Reflectance}
Although the reflectance layer remains invariant under different lighting conditions, the corresponding reflectance obtained in low-light scenarios is inevitably contaminated by complex noise. Consequently, the distribution of the low-light reflectance can be regarded as a perturbation of its normal-light counterpart, necessitating a powerful denoiser to remove complex noise. While denoising networks can be designed via end-to-end training, we find the rectified flow model used in illumination enhancement presents a compelling idea for data augmentation. Specifically, given noisy and clean reflectance pair $(R_l, R_n)$, one can generate more data pairs as $(R_t, R_n)$, with
\begin{equation}
\label{equ 8}
R_t = R_l + t(R_n-R_l), t \in [0,1],
\end{equation}
where $R_n \sim P(R_n)$ and $R_l \sim P(R_l)$.

Inspired by DnCNN \cite{zhang2017beyond}, our enhanced denoising network CRFR (as shown in Fig. \ref{fig1} and denoted by $v_t(R_t;\theta)$) is trained to predict the noise residual $(R_n - R_t)$ instead of directly regressing $R_n$. Thus, we can obtain an estimated clean reflectance as:
\begin{equation}
    \hat{R}_l = R_t +t v_t(R_t;\theta), \forall t \in [0,1].
\end{equation}

To train the enhanced denoising network module, we designed the following loss terms as follows:
\begin{align}
   {L_{\mathrm{content-R}}} &=   {1 - {\mathrm{SSIM}}\left( {{{\hat R}_l},{R_n}} \right)}, \label{equ9} \\
     {L_\mathrm{CRFR - C}} &= \left\| {{v_t(R_t;\theta)} - \left( {R_n - R_l} \right)} \right\|_F^2  +\left\| {v_t(R_t; \theta,2d) - {s_{\mathrm{target-R}}}} \right\|_F^2), \label{equ10} \\
    loss_2 &= {L_{\mathrm{content-R}}} +  {L_\mathrm{CRFR - C}}, \label{equ11}
 \end{align}
 where the prediction vector field $v_t(R_t; \theta) = \mathrm{CRFR_\theta}(R_t,t,d)$ with $d=0$, $s_\mathrm{target-R}=\frac{v_t(R_t; \theta, d)+v_t(R_{t+d}^{\prime}; \theta, d)}{2}$ and $R_{t+d}^{\prime} = R_t + dv_t(R_t; \theta,d)$. By combining data augmentation and consistency loss constraints, the enhanced denoising module can achieve superior denoising results compared to the general end-to-end denoising module.

\subsection{Inference Processes}
We begin by decomposing the input low-light image into its illumination $L_{l}$  and reflectance $R_{l}$ components, as illustrated by Fig. \ref{fig 2} the pre-processing part. 

Given the enhanced denoising network module (CRFR), we first estimate the noise residual $V_1 = R_n - R_l$ from the noisy input $R_l$ by computing $V_{1} = \mathrm{CRFR}_\theta (R_l,0,0)$. The denoised reflectance $\hat{R}_l$ can then be obtained as
\begin{equation}
	\label{equ17}
	\begin{split}
		\hat R_l = R_l + V_{1}.\\
		% \widehat{HF}_{low}^K = HF_{low}^K + V_{12}.
	\end{split}
\end{equation}

To refine the illumination component, we introduce two CRFI-based schemes: the first applies a one-step enhancement to a desired illumination level, while the second supports progressive (or continuous) adjustment across varying illumination intensities. Specifically, the one-step enhancement is written as
  \begin{equation}
  	\label{equ18}
  	\begin{split}
  		\hat L_l = L_l + \mathrm{CRFI}_{\theta}(L_l,0,0),
  	\end{split}
  \end{equation}
where the reflectance $\hat {L}_l$ follows the distribution of the normal-light reflectance component.
% The reconstructed illumination layer $\hat L_l$ corresponding to the low-light illumination layer $L_l$ can be obtained by a one-step forward Euler discritization. 
In contrast, the multi-step enhancement is written as
\begin{equation}
	\label{equ19}
	\begin{split}
		L_{t_{n+1}} = L_{t_n} + \frac{1}{N} \mathrm{CRFI}_{\theta}(L_{t_n},t_n,0),
	\end{split}
\end{equation}
 where the time interval $[0,T]$ is divided into $N$ segments, each with a length of $d$, and the first $n$ segments denote $t_n = n \times d$. Specifically, we set $T=1$ for experiments on the LOLv1 dataset. Note that $L_{t_0}$ is $L_l$ and $L_{t_N}$ is $\hat L_l$. From Eq. (\ref{equ19}), we obtain a sequence of illumination levels  $\left\{ L_{t_i} \right\}_{i=1}^N$.
 
 By fusing the denoised reflectance layer with either the single enhanced illumination ${\hat{L}}_l$ or the multi-level sequence $\left\{ L_{t_i} \right\}_{i=1}^N$, we achieve flexible, progressive enhancement of low-light images (Fig. \ref{fig 2} the restoration part).

\section{Experiments}
\subsection{Configuration, Dataset, and Methods}
\textbf{Implementation Details.}
The pretrained TDN architecture follows the Diff-Retinex \cite{yi2023diff}.  Both CRFR and CRFI modules are built on the SR3 backbone \cite{saharia2022image}, consisting of stacked residual blocks with attention mechanisms.
The CRFR is implemented in PyTorch and trained on two Tesla V100 GPUs. The input image size is set to $128 \times 128$, and the batch size is 32. Training is conducted for 100K iterations using the Adam optimizer \cite{kingma2014adam} with a learning rate of 0.0001.
Our CRFI is implemented in Pytorch on one Tesla-V100 GPU. The input image is of size $128 \times 128$, and the batch size is 20. Similarly, the Adam optimizer with a learning rate of 0.0001 is used to train the network for 300K iterations.
During training, half of each batch size is used to optimize the conditional rectified flow loss, and the other half is used to optimize the consistency constraint loss. The timestep $d$ is configured following the shortcut model \cite{frans2024one}.

 \begin{figure}[!ht]
	\centering
	\includegraphics[width=3.5in]{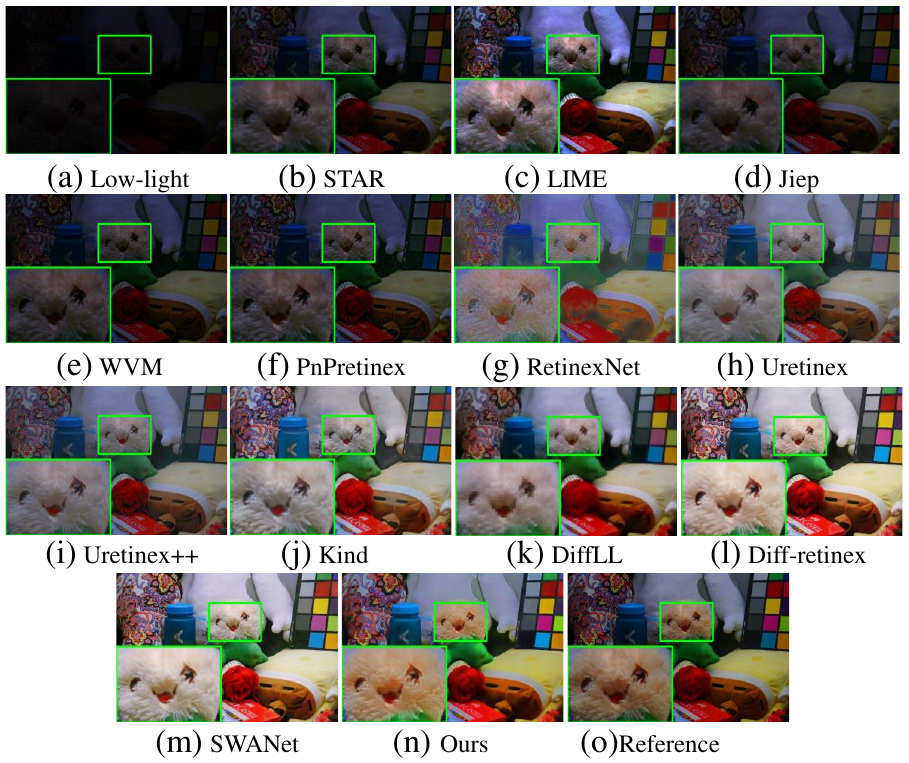}
	\caption{Qualitative comparison on LOLv1 dataset (Green boxes highlight brightness/color error-prone areas).}
	\label{fig 4}
\end{figure}

\textbf{Datasets.} IllumFlow is trained and evaluated on the LOLv1 dataset \cite{wei2018deep}. The LOL-v2 dataset \cite{yang2021sparse} is also used to assess the performance of our method. To further evaluate its generalization ability, we conduct experiments on the unpaired real-world benchmark dataset MEF \cite{ma2015perceptual}. The above datasets are publicly used in low-light enhancement.

\begin{figure}[!ht ]
	\centering
	\includegraphics[width=3.5in]{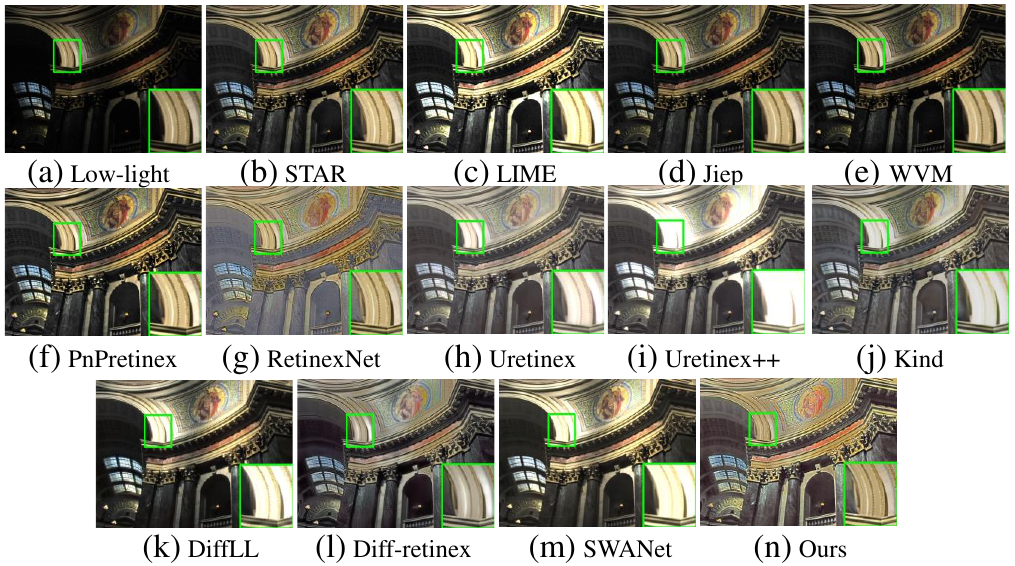}	
	\caption{Qualitative comparison on MEF dataset (green highlights indicate artifact-prone regions).}
	\label{fig 6}
    \end{figure}

\textbf{Comparison Methods.} The comparison methods are divided into three categories. Retinex-based optimization methods include STAR \cite{xu2020star}, LIME \cite{guo2016lime}, Jiep \cite{cai2017joint}, WVM \cite{fu2016weighted}, PnPretinex \cite{lin2022low}. Learning-based methods include RetinexNet \cite{wei2018deep}, Uretinex-net \cite{wu2022uretinex}, Uretinex++ \cite{wu2025interpretable}, KinD \cite{zhang2019kindling} and SWANet \cite{10244055}. Generative-based include DiffLL \cite{jiang2023low}, Diff-retinex \cite{yi2023diff}. 

 \subsection{Results and Analysis}
 \textbf{Low-Light Image Enhancement.}
We evaluate IllumFlow against state-of-the-art methods through comprehensive quantitative and qualitative comparisons. The reflectance layer is reconstructed using our enhanced denoising model (Eq. \ref{equ17}), while the illumination layer is restored via one-step enhancement (Eq. \ref{equ18}).

\begin{figure}[!ht]
	\centering
\includegraphics[width=0.5\textwidth]{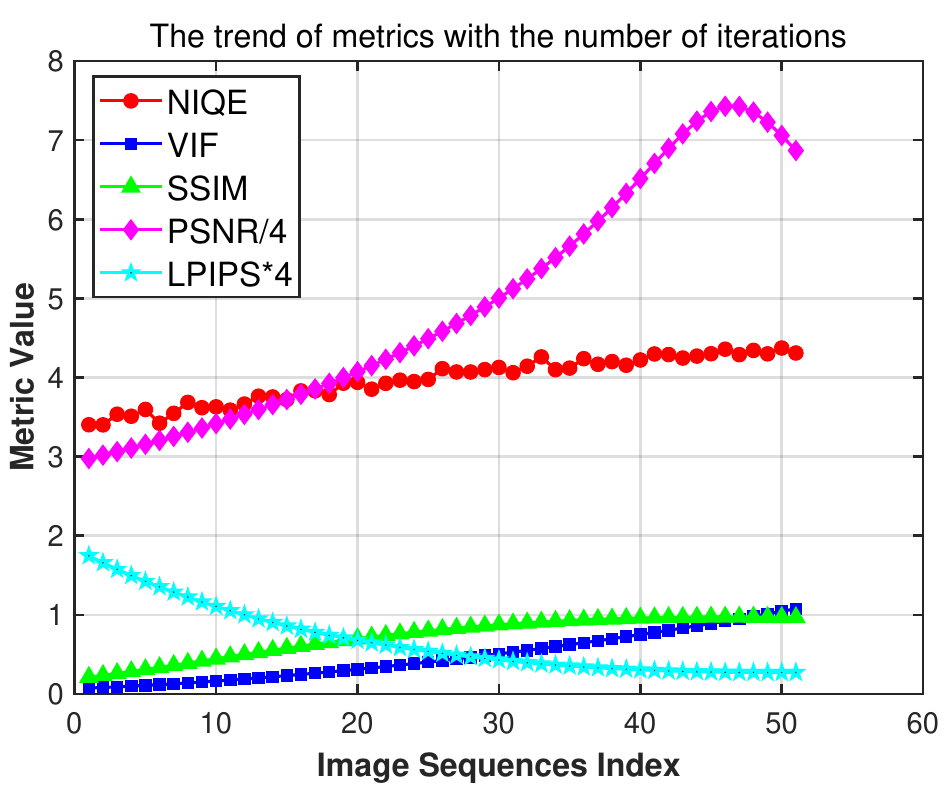}
	\caption{Metric trends: PSNR improves with higher illumination levels but shows slight degradation as iterations approach 45, NIQE increases steadily, while SSIM and LPIPS converge by ~40 iterations.}
	\label{fig 8}
\end{figure}

For quantitative evaluation, we adopt both reference-based metrics - PSNR, SSIM \cite{wang2004image}, and LPIPS \cite{zhang2018unreasonable} - as well as no-reference metrics - NIQE \cite{mittal2012making} and VIF \cite{sheikh2005visual}. 
The results are shown in the Table. \ref{table 1}.
It can be seen that our method achieves outstanding performance in reference-based metrics PSNR, SSIM and LPIPS on the LOLv1 dataset and LOLv2 real-captured.
Our method demonstrates superior performance on both NIQE and VIF metrics for the unpaired MEF dataset.
Qualitative comparisons in Figs. \ref{fig 4} and \ref{fig 6} demonstrate our method's superior visual performance, particularly in: noise suppression, color fidelity preservation, natural brightness maintenance. Our results show significant perceptual improvements over competing approaches.

% \begin{sidewaystable}[ht]
% \begin{turn}{90}
\begin{table*}[!ht]
 	\centering
     \resizebox{1\textwidth}{!}{
 	\begin{tabular}{c|c|ccc|ccc|cc}
 		\toprule
 		\multirow{2}{*}{Method} & \multirow{2}{*}{Reference} & \multicolumn{3}{|c|}{LOLv1} & \multicolumn{3}{|c|}{LOLv2 real-captured} & \multicolumn{2}{|c}{MEF}\\
 		\cline{3-10}
 		~ & ~ &  PSNR $\uparrow$ & SSIM $\uparrow$ & LPIPS $\downarrow$ & PSNR $\uparrow$ & SSIM $\uparrow$ & LPIPS $\downarrow$ & NIQE $\downarrow$ & VIF $\uparrow$ \\
% 		Methods & Reference &  LOL & LOLv2 real-captured & DICM
% 		Dataset & ours & STAR & dong & MF & LIME & JIep & WVM & Li & PnPRetinex \\
 		\midrule
 		STAR & TIP'20 & 12.6386 & 0.5375 & 0.3097  & 15.5757 & 0.5718 & 0.2760 & \underline{3.1401} & 1.6050 \\
 		LIME & TIP'17 & 16.9204 & 0.5990 & 0.3605 & 17.7806 & 0.5455 & 0.3528 & 3.5438 & \textbf{3.8041} \\
 		Jiep & ICCV'17 & 12.0466 & 0.5124 & 0.3157  & 14.7192 & 0.5884 & 0.2804 & 3.1616 & 1.8578 \\
 		WVM & CVPR'16 & 11.8552 & 0.4979 & 0.3401  & 14.4505 & 0.5421 & 0.3117 & 3.2041 & 1.8984 \\
 		PnPretinex & TIP'22 & 13.0721 & 0.5775  & 0.6953 & 16.1438 & 0.5826 & 0.3169 & 3.1651 & 2.9933 \\
 		RetinexNet & BMVC'18 & 18.8685 & 0.6943 & 0.3863  & 17.8736 & 0.6648 & 0.4388 & 3.6185 & 1.5215 \\
 		Uretinex & CVPR'22 & 21.0536 & 0.8306 & 0.3847  & 20.4114 & 0.8547 & 0.2315 & 3.3726 & 1.5221 \\
 		Uretinex++ & TPAMI'25 & 23.0251 & 0.8395 & 0.2981  & 24.1302 & 0.8601 & 0.2113 & 3.4899 & 2.0006 \\
 		KinD & MM'19 & 19.6554 & 0.8214 &  0.1558  & 21.1405 & 0.8550 & 0.1412 & 3.7042 & 1.6711 \\
 		DiffLL & TOG'23 & \underline{26.3342} & 0.8447 & 0.1184 & 28.8522  & 0.8746 & 0.0999 & 3.4275 & 2.2721 \\
 		Diff-retinex & ICCV'23 & 22.7118 & 0.8551 & 0.1997  & 26.6117 & 0.9021 & 0.1797 & 3.2185 & 1.8017 \\
        SWANet & TCSVT'24 & 25.3725 & \underline{0.8596} & \underline{0.1167}  & \textbf{30.6853} & \underline{0.9118} & \underline{0.0793} & 3.5127 & 2.4851 \\
 		ours & - & \textbf{26.6260} & \textbf{0.8739} & \textbf{0.0769} & \underline{28.8907}& \textbf{0.9129} &\textbf{0.0752} & \textbf{3.0191} & \underline{3.2150} \\
 		\bottomrule
 	\end{tabular}
    }
    \caption{Quantitative results of different low-light image enhancement methods on the LOLv1, LOLv2 real-captured paired datasets and MEF unpaired dataset. The best results are highlighted in \textbf{bold} and the second best results are \underline{underlined}.}
 	\label{table 1}
 \end{table*}
 % \end{sidewaystable}

Note that a sequence of illumination-enhanced results can also be generated via the multi-step forward Euler method Eq. (\ref{equ19}), enabling progressive enhancement at varying illumination levels. Fig. \ref{fig 8} shows metric trends versus illumination level (vertex: illumination intensity; horizontal axis: metric values). It can be seen from Fig. \ref{fig 8} metrics peaking at illumination index 46, with optimal quality at $t\in [0,1]$ (vertical index 51).

\begin{figure}[!ht ]
	\centering
    \includegraphics[width=4.3in]{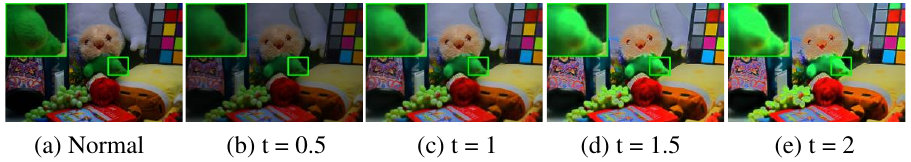}
\\
	\caption{ Illumination improves progressively with $t \in [0,2]$, enhancing brightness and color fidelity.}
	\label{fig 9}
\end{figure}

\begin{figure}[H]
	\centering
    \includegraphics[width=3.4in]{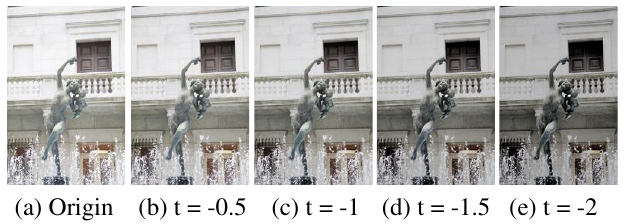}\\
	\caption{Illumination suppression. As time $t$ decreases, the illumination contrast is improved.}
	\label{fig 10}
\end{figure}

\textbf{Exposure Correction.}
Our IllumFlow leverages a bidirectional ODE to enable flexible illumination control: a) Enhancement ($t>0$): Boosts brightness/contrast (Fig. \ref{fig 9}); b) Suppression ($t<0$): Reduces illumination (Fig. \ref{fig 10}).
This unified framework supports arbitrary adjustment levels through a single parameter $t$. 

By simply adjusting the parameter $t$, IllumFlow dynamically controls illumination levels without requiring additional network modules, effectively compensating for performance degradation caused by low-quality training samples.

\begin{figure}[H]
	\centering
    \includegraphics[width=4in]{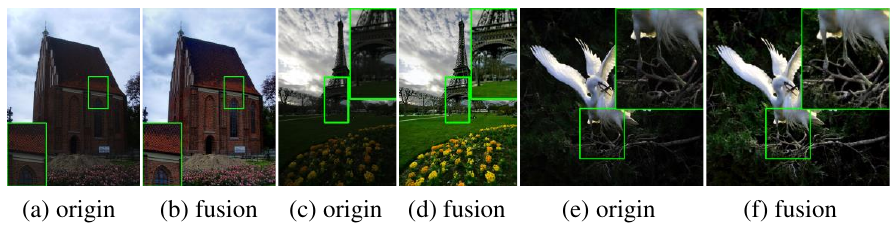}\\
	\caption{Synthesize high-quality images from multi-exposure sequences generated by our method.}
	\label{fig 11-1}
\end{figure}

Additionally, our method enables exposure correction via single-image multi-exposure generation, compatible with standard MEF techniques, such as the fast multi-scale MEF (FMMEF) approach \cite{li2020fast}, to produce high-quality fused results (Fig. \ref{fig 11-1}). More results please refer to the appendix.

\textbf{Ablation Study.}
To evaluate our key modules, we design five ablation variants:
1) Full end-to-end: Both reflectance and illumination are learned directly. 2) Hybrid-Ret: End-to-end reflectance + CRF-based illumination; 3) Diff-Ret: Diffusion-based reflectance + CRF illumination; 4) Consist-Ret: Diffusion with consistency refinement + CRF illumination; 5) Full-CRF: CRF-regularized reflectance and illumination.
% In these ablation settings, the network architecture remains consistent with that used in our main experimental framework. 
We evaluate our method on LOLv1 and LOLv2 datasets using PSNR and SSIM metrics. As Table \ref{table 3} shows, Setting 5 (Full-CRF) achieves optimal performance. For fair comparison in Setting 3 (Diff-Ret), we remove the consistency network from Diff-Retinex  \cite{yi2023diff} and retrain the diffusion model.

\section{Conclusion}
In this paper, we propose a novel framework, IllumFlow, to address two key challenges in LLIE: illumination variations and complex noise. Our method first decomposes low-light images into reflectance and illumination layers. Then, we employ a conditional rectified flow model to create a continuous flow field for the illumination layer to adapt to a wider range of lighting conditions, and propose an enhanced denoiser by data augmentation for the reflectance layer. Our approach enables bidirectional and continuous linear illumination control. Extensive experiments validate the superior performance and flexibility of our method compared to existing approaches.

\begin{table}[H]
	\centering
     \resizebox{0.5\textwidth}{!}{
	\begin{tabular}{c|cc|cc}
		\toprule
        \multirow{2}{*}{Setting} & \multicolumn{2}{|c|}{LOLv1} & \multicolumn{2}{|c}{LOLv2 }\\
        \cline{2-5}
        ~  &  PSNR $\uparrow$ & SSIM $\uparrow$  & PSNR $\uparrow$ & SSIM $\uparrow$ \\
		% \diagbox{Setting}{Metric} & PSNR $\downarrow$ & SSIM $\uparrow$ & PSNR $\downarrow$ & SSIM \\
		\midrule
		1 & 23.3352  & 0.8512    & 24.4632 &0.8815   \\
		2 & 25.8259  & 0.8702    & 27.4321 & 0.8987    \\
		3 & 15.9961  & 0.7108    & 18.2973 &0.7607   \\
            4 &  23.6008    & 0.8678 & 27.5379 & 0.9113\\
		5 & 26.6260	& 0.8739	&28.8907 &0.9129 \\
		\bottomrule
	\end{tabular}
    }
    \caption{The metrics for several network settings.}
	\label{table 3}  % 移除空格
\end{table}

\section{Acknowledgments}
The authors thank the editors and the anonymous reviewers for their constructive comments and suggestions. This paper is supported by the National Natural Science Foundation of China (Grants Nos. 62372359, 61772389, 61972264) and the Xidian University Specially Funded Project for Interdisciplinary Exploration (TZJHF202513).

\section{Appendix}

\begin{figure}[!ht ]
	\centering
	\includegraphics[width=3.2in]{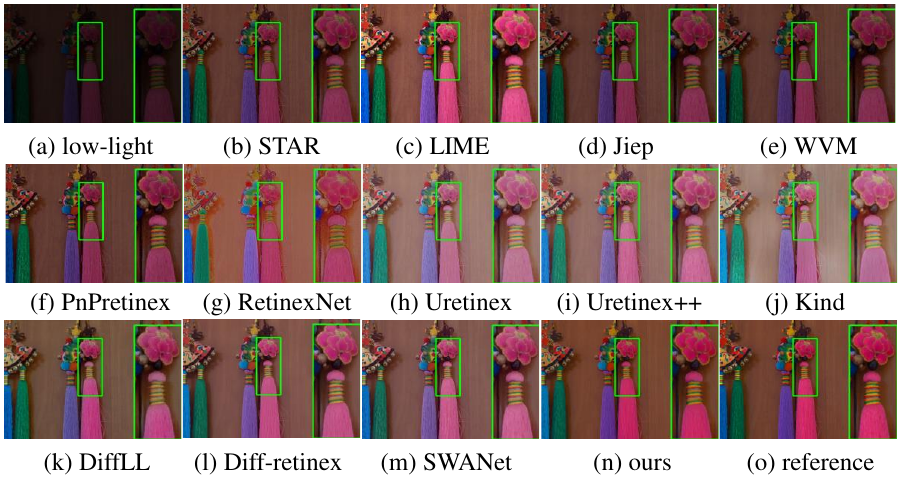}
	\\	
	\caption{Results on LOLv1 dataset with different methods (Green boxes highlight brightness/color error-prone areas).}
	\label{fig 2_app}
\end{figure}
\subsection{Supplement for LLIE}

\begin{figure}[!ht ]
	\centering
	\includegraphics[width=3.2in]{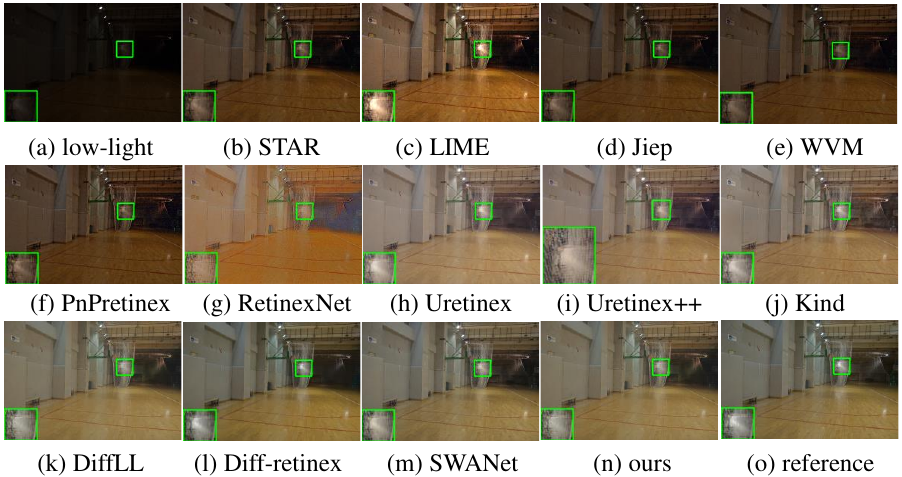}
	\\	
	\caption{Results on LOLv2 real-captured dataset with different methods (Green boxes highlight brightness/color error-prone areas).}
	\label{fig 5_app}
\end{figure}

\begin{figure}[ht ]
	\centering
	\includegraphics[width=3.2in]{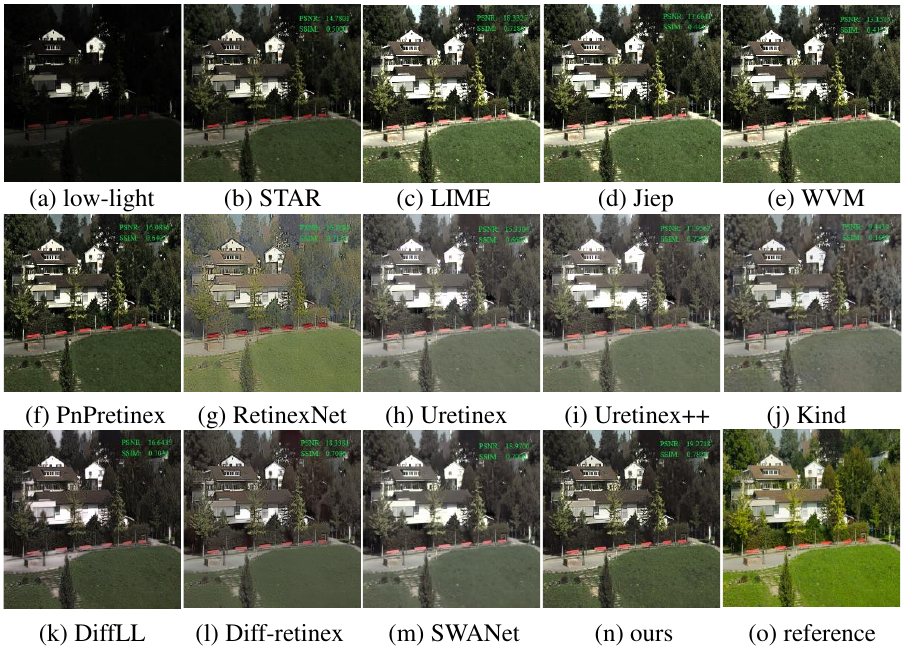}
	\\	
	\caption{Results on LOLv2-Synthetic dataset with different methods. The metrics of the results produced by comparison methods are marked in green, and our method has the highest performance.}
	\label{fig 6_app}
\end{figure}

\begin{figure*}[!ht ]
	\centering
	\includegraphics[width=3.2in]{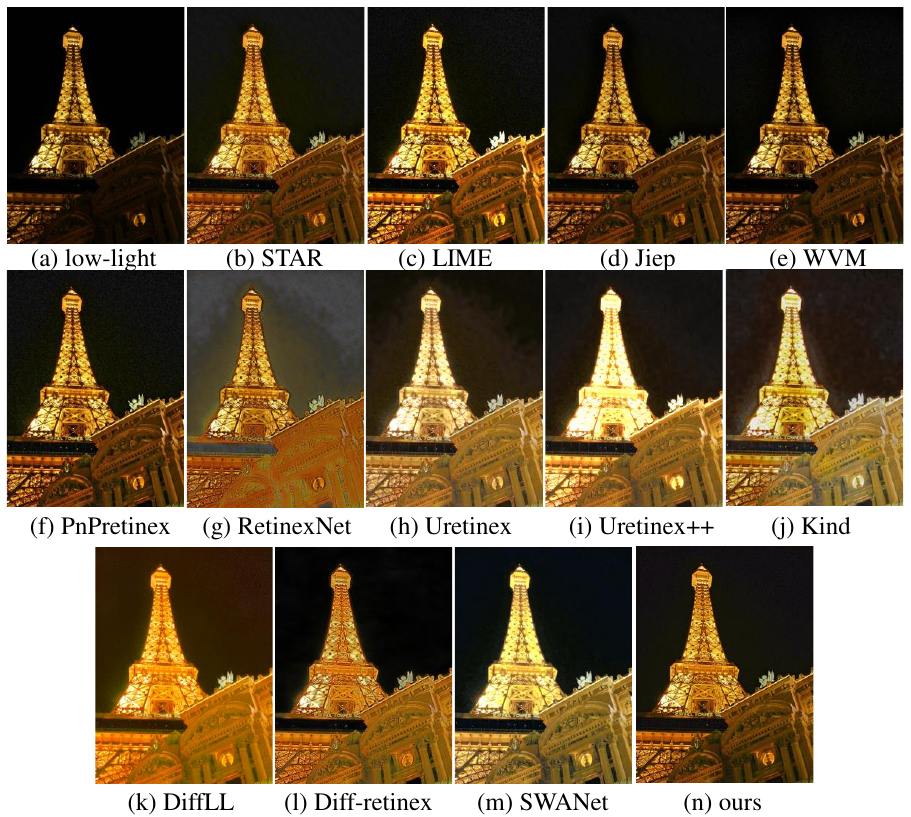}
	\\	
	\caption{Results on DICM \cite{lee2013contrast} dataset with different methods.}
	\label{fig 7_1_app}
\end{figure*}

\begin{figure*}[!ht ]
	\centering
	\includegraphics[width=3.2in]{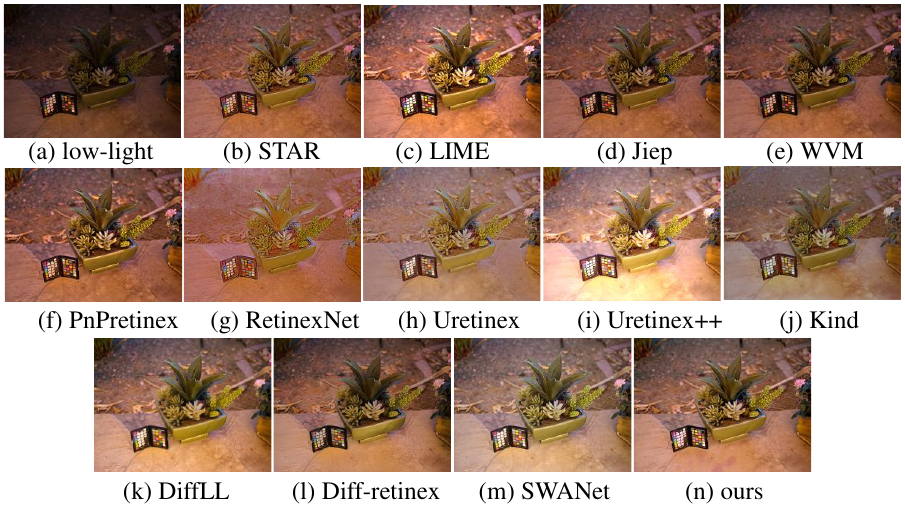}
	\\	
	\caption{Results on LIME \cite{guo2016lime} dataset with different methods.}
	\label{fig 7_2_app}
\end{figure*}

We provide additional qualitative results to demonstrate the effectiveness of our method in low-light image enhancement. Fig. \ref{fig 2_app} presents a visual comparison between our approach and other state-of-the-art methods on the LOLv1 dataset. Fig. \ref{fig 5_app} and Fig. \ref{fig 6_app} illustrate the comparisons on the LOLv2 real-captured and synthetic datasets, respectively. 
To further validate the generalization ability of our method, we evaluated its performance on two unpaired datasets, DICM and LIME, using the NIQE metric. The quantitative results are summarized in the Table. \ref{table 1_app}. In addition, qualitative comparisons on the DICM and LIME datasets are provided in Fig. \ref{fig 7_1_app} and Fig. \ref{fig 7_2_app}, respectively. In general, our method achieves superior performance both quantitatively and qualitatively across diverse datasets.

 \begin{table}[H]
 	\centering
    \resizebox{0.5\textwidth}{!}{
 	\begin{tabular}{c|c|c|c}
 		\toprule
 		\multirow{2}{*}{Method} & \multirow{2}{*}{Reference} & \multicolumn{1}{|c|}{DICM} & \multicolumn{1}{|c}{LIME} \\
 		\cline{3-4}
 		~ & ~ &  NIQE $\downarrow$ & NIQE $\downarrow$ \\
% 		Methods & Reference &  LOL & LOLv2 real-captured & DICM
% 		Dataset & ours & STAR & dong & MF & LIME & JIep & WVM & Li & PnPRetinex \\
 		\midrule
 		STAR & TIP'20 & 4.4596 & 4.6921  \\
 		LIME & TIP'17 & 3.5772 & \textbf{4.1377}  \\
 		Jiep & ICCV'17 & 4.9256 & 4.5982  \\
 		WVM & CVPR'16 & 4.7039 & 4.5149\\
 		PnPretinex & TIP'22 & 4.0149& 4.4764  \\
 		RetinexNet & BMVC'18 & 4.6364 & 4.8279  \\
 		Uretinex & CVPR'22 & 3.6537 & 4.7738  \\
 		Uretinex++ & TPAMI'25 & 3.5278 & 4.8475 \\
 		KinD & MM'19 & 4.2364 & 5.7548  \\
 		DiffLL & TOG'23 & 3.5172& 4.2399  \\
 		Diff-retinex & ICCV'23 & \underline{3.4120} & 4.
        7387\\
        SWANet & TCSVT'24 & 3.8195 & 5.
        3163\\
 		ours & - & \textbf{3.3556} & \underline{4.1650} \\
 		\bottomrule
 	\end{tabular}
    }
    \caption{Quantitative results of different low-light image enhancement methods on the DICM and LIME unpaired dataset. The best results are highlighted in \textbf{bold} and the second best results are \underline{underlined}.}
 	\label{table 1_app}
 \end{table}

\subsection{Supplementary for Fig. 8.} 
\begin{figure}[H]
	\centering
	\includegraphics[width=0.6\textwidth]{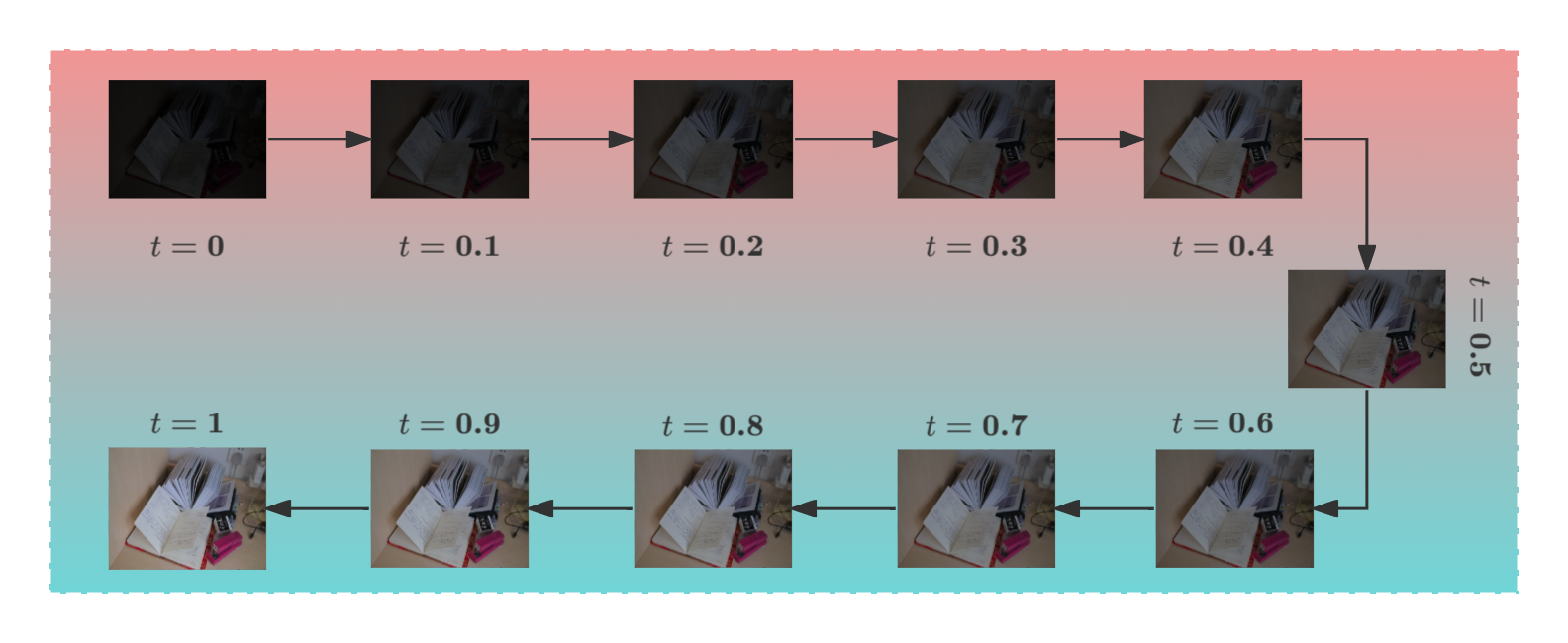}
	\caption{Illumination variation of image sequences governed by a forward ODE with 50-step discretization.}
	\label{fig 7_app}
\end{figure}

Our methods attain illumination enhancement sequences through a multi-step forward Euler method, which  enables continuous illumination control,  which are illustrated in Fig. \ref{fig 7_app}. As can be seen from Fig. \ref{fig 7_app}, the visual effect of the image sequence gradually increases. The brightness level can be adjusted by adjusting the time $t \in [0,1]$.
It is a distinctive advantage over previous methods.

\subsection{CLE-diffusion VS IllumFlow.}

The CLE-Diffusion method \cite{yin2023cle} can also achieve the goal of controllable light levels. This approach enables users to control the desired illumination levels by introducing illumination embeddings. However, this method can only adjust a single illumination level per inference. Additionally, the diffusion-based framework it employs results in prolonged inference times. Implementing continuous control of lighting embeddings would incur substantial computational costs.
\begin{figure}[H]
	\centering
	\includegraphics[width=2.8in]{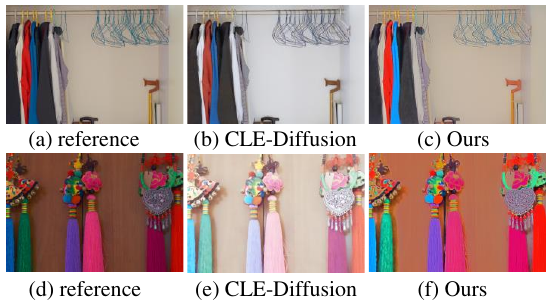}
	\\
	\caption{Comparative Analysis of Over-Enhancement Effects: CLE-Diffusion vs. Our Method}
	\label{fig 11_app}
\end{figure}

In contrast, our approach achieves continuous lighting control during a single inference by leveraging the discrete iteration process of ODE. The comparison of inference speed between our method and CLE-Diffusion is presented in the Table. \ref{table 2}.
 \begin{table}[!htbp]
 	\centering
    \resizebox{0.5\textwidth}{!}{
 	\begin{tabular}{c|c|c|c}
 		\toprule
 		Metric & CLE-Diffusion & Ours & Ours \\
 		\midrule
 		Time (s/infer) & 63.7953 & 0.0517 & 11.8323 \\
 		Number(n/infer) & 1 & 1 & 50 \\
 		\bottomrule
 	\end{tabular}
    }
    \caption{Comparison of inference speed between our method and CLE-Diffusion. Our method demonstrates efficient inference capabilities}
 	\label{table 2}  % 移除空格
 \end{table}
Our method requires significantly less inference time to produce a single enhanced result compared to CLE-Diffusion as shown in the Table. \ref{table 2}. Remarkably, our approach can even generate 50 continuous illumination image sequences in less time than CLE-Diffusion takes to produce just one enhanced image. Furthermore, qualitative evaluation of over-enhanced results demonstrates that our method better preserves color consistency compared to CLE-Diffusion, as evidenced in Fig. \ref{fig 11_app}.

\subsection{Supplement for Exposure correction.}

\begin{figure}[H]
	\centering
	\includegraphics[width=3in]{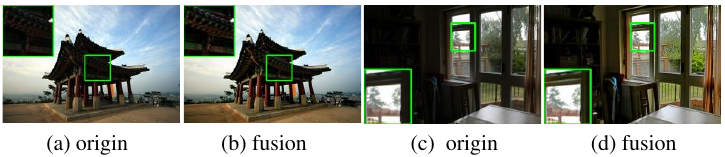}
	\\
	\caption{Synthesize high-quality images from multi-exposure sequences generated by our method.}
	\label{fig 12_app}
\end{figure}

\begin{figure}[H]
	\centering
	\includegraphics[width=3in]{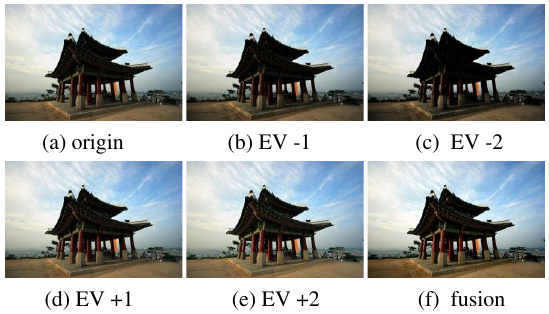}
	\\
	\caption{Generated static multi-exposure sequences by our method.}
	\label{fig 13_app}
\end{figure}
Leveraging IllumFlow, the pretrained model can be directly applied to the multi-exposure image fusion (MEF) algorithm \cite{zhang2021benchmarking} to improve image quality. 
In practice, acquiring multi-frame images inevitably introduces motion—whether due to camera shake or object movement \cite{chen2022attention}—which necessitates the design of alignment modules. 
This significantly increases the complexity of the method design. Moreover, publicly available static multi-frame datasets are rare, making multi-frame fusion even more challenging.

Our approach effectively addresses this issue by generating multi-exposure image sequences from any single low-quality image, without requiring any specialized datasets. Since the generated sequences are inherently static, they eliminate the need for alignment networks and allow the direct use of existing static multi-frame fusion methods, such as fast multi-scale MEF (FMMEF) \cite{li2020fast}, to produce a single visually pleasing and high-quality fused image.

We apply continuous exposure adjustment to a set of images generated by our method and perform high-quality fusion using the FMMEF algorithm. The more results, as shown in Fig.\ref{fig 12_app}, demonstrate the effectiveness of our approach in producing a high-quality fused image.

The Fig. \ref{fig 13_app} shows a static multi-exposure sequence generated by ILLumFlow based on Fig. \ref{fig 12_app} (a) to attain the fusion image Fig. \ref{fig 12_app} (b). We implement multi-exposure sequences by adjusting time $t$ through a bidirectional ODE. Note that $EV \pm N$ is the simulated exposure $\pm N$ stop.

\bibliographystyle{elsarticle-num} 
\bibliography{references}
\end{document}